# Advancements in Gesture Recognition Techniques and Machine Learning for Enhanced Human-Robot Interaction: A Comprehensive Review

[1]Sajjad Hussain[*], [1]Khizer Saeed, [1]Almas Baimagambetov, [1]Shanay Rab and [2]Md Saad

*Multidisciplinary Robotics, Control and Artificial Intelligence (MOCAI) Research Laboratory*

*School of Architecture, Technology and Engineering, University of Brighton, Brighton, UK*

[2]*Department of Mechanical Engineering, Jamia Millia Islamia, New Delhi, India*

[*]*Corresponding author email: s.hussain4@brighton.ac.uk*

**Abstract:**

In recent years robots have become an important part of our day-to-day lives with various applications. Human-robot interaction creates a positive impact in the field of robotics to interact and communicate with the robots. Gesture recognition techniques combined with machine learning algorithms have shown remarkable progress in recent years, particularly in human-robot interaction (HRI). This paper comprehensively reviews the latest advancements in gesture recognition methods and their integration with machine learning approaches to enhance HRI. Furthermore, this paper represents the vision-based gesture recognition for safe and reliable human-robot-interaction with a depth-sensing system, analyses the role of machine learning algorithms such as deep learning, reinforcement learning, and transfer learning in improving the accuracy and robustness of gesture recognition systems for effective communication between humans and robots.

Keywords: Gesture recognition; Machine learning; human-robot–interaction,

## Introduction:

Human-robot interaction (HRI) is the fastest-growing area in robotics with the integration of artificial intelligence and machine learning. Gesture recognition is the way to communicate with robots. Robots are fast and very accurate machines to perform various tasks. Robotic systems provide better accuracy, flexibility, and productivity. Human-robot interaction is the key to establishing the communication channels between humans and robots. This paper provides a comprehensive review of the advancement in the gesture recognition technique and the different algorithms used to enhance gesture recognition for safe and reliable human-robot interaction. In daily life, gestures play a crucial role. Vision-based gesture recognition is an approach that combines computer pattern recognition with advanced sense. It is essential to enhance human–robot–interaction and is employed in a wide range of application engineering and research. The current state of gesture detection technology falls short of achieving real-time human–robot interaction because natural gestures are always changing as well as the complex background in gesture recognition [1].

The focus of human-robot interaction is on the interaction between humans and their machines and developing robotic systems capable of understanding and responding to human behavior and common tasks. The robot must be able to sense, understand, and react to human states, actions, and emotions to behave in ways that are acceptable to society. Using the person's visual perception, robot social sensitivity can be improved [2]. This is connected to human-robot interaction. Instead, human-robot collaboration emphasizes how people and machines cooperate to accomplish common objectives with a purpose and expected result. In HRC, robots assist humans in achieving their intended goals or contribute value to them [3]. Collaborating with a robot can enhance task speed and efficiency, reduce errors, and enhance human safety to prevent tiredness and injuries [3].

The motivation behind this review is rapid advancements in both gesture recognition with the help of different machine learning and deep learning algorithms. By synthesizing and analyzing the latest research findings, methodologies, and applications, this paper seeks to offer insights into the current trends, challenges, and future directions in the field of enhanced human-robot interaction through vision-based gesture recognition. The structured examination of gesture recognition techniques, machine learning approaches, applications in robotics, and key challenges, this review aims to provide a valuable resource for researchers, engineers, and practitioners working at the intersection of robotics, AI, and human-computer interaction. By understanding the intricacies and potential of gesture-based communication, we can push the way for more safe, efficient, and human-centric interactions between humans and robots.

## Gesture Recognition Techniques:

Vision-based gesture recognition techniques involve using cameras to capture and interpret human gestures. Vision-based gesture recognition is the nonverbal type of gesture recognition technique it establishes the communication channel between the human and robot by capturing the physical movement or pose provided by the human, including the movement of the hand, arm, body, and face which is meaningful and informative. There are different techniques including deep learning methods like convolutional neural networks (CNNs) and traditional approaches like Hidden Markov Models (HMMs) and Support Vector Machines (SVMs). They analyze images or video frames to identify and classify gestures accurately, enabling intuitive human-robot interaction. This paper explores the effectiveness of these vision-based techniques in enhancing gesture recognition accuracy for improved human-robot communication [4]. For modern robots to understand and communicate with the actual world more effectively and to make relevant decisions and choices based on visual data, computer vision is necessary [5]. RGB (red, green, and blue wavelength) cameras are common sensors used in computer vision; they capture light in RGB and convert it into a color representation of the environment to offer detailed information. Robots and humans can communicate details, objects, and task-related information by using visual information to interpret the environment, which can assist mimic how humans see the world. Computer vision utilizes methods including pixel-level classification to identify which region of an image is part of an area of interest, object detection to determine the location of an object in the picture, and image classification to identify what's in the image [5]. When it comes to computer vision for individuals, techniques like face detection [6], posture estimation [7], and human motion tracking [8] can be used to recognize and analyze humans in visual settings. Subsequently, this kind of visual data can help create mutual understanding and knowledge between humans and robots. Between 2010 and 2020, there has been a significant evolution in the field of computer vision. Since it won the 2012 ImageNet competition, deep learning has become increasingly important [9]. Gesture recognition has been tested as a communication and control system, converting human poses into command signals to activate state changes or initiate information flow between humans and robots [10]. Additionally, gestures can be utilized to tell a robot where to move [11] and what object to use [12]. Through perception and interpretation of individuals and items in the scene, including human grasp ability, mobility, and collision range, visual information can also facilitate collaborative robot actions with the person, such as human-robot object exchange[13]. The field of computer vision offers a wealth of opportunities to enhance robot perception and action based on visual input, hence encouraging better human-robot collaboration. To enhance robot functioning, user experience, interface design, control strategies, and robot utility for certain actions or tasks, this includes robotic vision.

Gesture recognition has two types [14], static gesture recognition and dynamic gesture recognition [15]. Static gesture recognition identifies the hand posture, shape, and position while dynamic gesture recognition identifies the composed of the sequential frames of the static gesture it recognizes the movement of the gesture [16].

The process of vision-based gesture recognition is as follows.

**Data acquisition:** Data acquisition is the process of capturing an image or video through the camera and preprocessing the image and video.

**Gesture Detection and Segmentation:** Gesture detection is the process of detecting an image and segmentation is used to segment the portion of the image and video.

**Gesture recognition:** obtaining information from the hand region of the image, then using the data to identify the type of gesture.

Vision-based gesture recognition technology refers to the acquisition of video images with operation gestures through acquisition devices such as cameras and the corresponding processing of video images, such as gesture segmentation, gesture feature extraction, gesture feature classification, etc. This technology was created to facilitate information exchange between users and intelligent devices. Table 1 summarizes key image processing algorithms typically employed in gesture recognition systems. These methods include edge detection to detect sharp borders, morphological processing to refine gesture shapes, optimal thresholding for precise segmentation, image smoothing to eliminate noise, and image grayscale to simplify processing. Each method is essential to enhance the precision and dependability of gesture recognition algorithms, which promotes more reliable and effective human-robot interaction.

*Table 1. Summary of Image Processing Techniques for Gesture Recognition*

| Image Processing Technique | Description | Reference |
|---|---|---|
| Image Grayscale | Conversion of color images to grayscale for simplified processing and reduced computational requirements. Methods include average, weighted average, maximum value, and minimum value. | [17-19] |
| Image Smoothing | Removal of noise using techniques like Gaussian filtering and median filtering. Gaussian filtering preserves edge information, while median filtering eliminates outliers. | [20-23] |
| Edge Detection | Identification of sharp discontinuities in an image using algorithms like Canny and Sobel. Canny algorithm removes noise effectively and extracts clear edges but is computationally intensive. Sobel algorithm is computationally simple and fast but may struggle with straight edges and noise. | [24-26] |
| Morphological Processing | Operations like expansion, erosion, opening, and closing to extract shape information of gestures. Expansion makes target regions more visible, erosion removes noise, and open/close operations refine gesture shapes. | [27] |
| Optimum Thresholding | Segmentation algorithms such as Otsu and Niblack divide images into foreground and background for better gesture segmentation. Otsu algorithm minimizes variances for adaptive thresholding, while Niblack algorithm handles diverse gray value distributions and uneven illumination effectively. | [28-30] |

For the machine to identify the gesture, it must first be distinguished from the background. The computer captures information about the gesture as well as the context in which it takes place, which illustrates it. In order to decrease the computation of pixel points and make the next actions easier, hand segmentation is the split of the collection of pixel point coordinates collected in the earlier gesture detection phase. Gesture segmentation is the first critical step in the algorithm for identifying gestures, and proper gesture identification depends on the successful completion of this stage. There are numerous methods for segmenting gestures, but practically all of them still encounter significant challenges in terms of accuracy, stability, and speed (for example, when segmenting gestures in complex backgrounds), the effect of the person's distance from the camera on the segmentation process, etc.

Gesture segmentation plays a crucial role in extracting meaningful gestures from images or video streams, enabling accurate interpretation and interaction in human-computer interfaces. Table 2 provides a comprehensive overview of various gesture segmentation techniques employed in digital systems. Contour information segmentation utilizing edge detection operators and template matching, and other segmentation approaches focusing on appearance features and deep learning methodologies. Each segmentation method is discussed in terms of its methodology, advantages, and references to relevant studies, offering insights into the diverse strategies used to achieve effective gesture segmentation in different contexts.

*Table 2. Summary of Gesture Segmentation Techniques*

| Summary of Gesture Segmentation Techniques | Description | Reference |
| --- | --- | --- |
| Skin Color Segmentation | Utilizes skin tone information for gesture segmentation, with RGB, HSV, and YCbCr color spaces commonly used. Various thresholding and skin color detection methods are applied for accurate segmentation. | [31-41] |
| Contour Information Segmentation | Detects gesture contours and edges using edge detection operators, template matching, and active contour models. Techniques such as histogram thresholding, morphological filtering, and wrist cropping are employed for segmentation. | [42-47] |
| Other Segmentation Approaches | Includes methods based on appearance features apart from skin color and contour, such as shape and direction analysis, adaptive threshold binarization, and depth sensor-based segmentation. Deep learning approaches are also discussed for automatic gesture segmentation. | [48-53] |

The most important stage in gesture recognition is gesture feature extraction. Processing the input gesture image and then extracting the features that can best represent the gesture from the image constitute the feature extraction step. The features that can be retrieved from the study of hand movements and postures to characterize the gesture shape and motion state are known as gesture features [54]. The selection and arrangement of gesture features affects the system's complexity and real-time performance in addition to its ability to recognize gestures accurately. The two primary categories of gesture features are global and local features.

**Global features:**

Global features define the complete hand's morphology and movement. These consist of the hand's dimensions, form, direction, speed, acceleration, rotation angle, etc. These features are relevant to certain basic gesture identification tasks and can characterize the hand's overall action state. Additionally common global features are Gabor filters, color and grayscale histograms, and more. A color histogram is a representation of the statistics of the frequency of occurrence of different hues in an image, shown as a histogram. A gray histogram, shown as a histogram, is a tally of how frequently each gray level appears in an image. The direction and frequency of an image's texture are detected by a Gabor filter, which is able to extract texture information from images and express it as a feature vector [54]. Global features offer high invariance properties, understandable representations, and are easy to compute. However, the majority of these features are represented by pixel-based point features, which leads to issues like high feature complexity and significant computational overhead. Furthermore, these feature descriptions don't work when there is image blending or occlusion.

**Local features:**

Local features are those obtained by studying specific locations, such as fingers, palms, and wrists. These consist of the wrist's rotation angle, the degree of palm protrusion, the curvature of the fingers, etc. These features are more suitable for gesture positions requiring high-precision recognition and can more properly characterize the detailed information about the hand. The shift-invariant feature transform (SIFT), the local binary pattern (LBP), the histogram of the oriented gradient (HOG), the sped-up robust features (SURF), features derived from principal component analysis (PCA), and linear discriminant analysis (LDA) are examples of frequently used local features.

Bill Triggs et al proposed the feature descriptor HOG in 2005 [55]. It is a statistical number that is used to describe features for target detection. In computer vision and image processing, determine the directional information of local picture gradients [55, 56]. The Histogram of Oriented Gradients (HOG) and SIFT methods are rated at a lower but still impressive rate of 91%, followed by PCA at a close rate of 91.5% which are investigated by the Gupta B et al [57] and Huong TNT et al [58]. A feature extraction technique used in computer vision and image processing is called LBP. By comparing the pixel's gray value's magnitude with that of the surrounding pixels, it transforms a pixel point into binary information and produces a feature that characterizes the image's texture. Features of LBP, like rotation invariance and grayscale invariance, offer important benefits [56, 59].

Ahmed AA, Aly S [60] proposed Local Binary Patterns (LBP) and Principal Component Analysis (PCA) methods have the highest accuracy at 99.97%. The Speeded Up Robust Features (SURF) method [61] has an accuracy of 63% and SIFT, Hu moments, and LBP are combined [97], the accuracy ranges between 87.3% and 85.1%.Baranwal N, Nandi GC [62] proposed the SURF method appears with the longest common subsequence feature, attaining an accuracy of 93%. Proposed by Lowe [63], SIFT is a scale- and rotation-invariant feature extraction technique that has found widespread application in gesture detection. It is an image processing approach for feature recognition and description that can identify and characterize feature points in images at various rotation degrees and sizes. SURF is a descriptor that was developed from SIFT that uses methods like integral image and quick Hessian matrix computation to increase the resilience and speed of feature recognition and description. When contrasted to SIFT, SURF has the clear benefit of being computationally quick [64]. With recognition rates of 81.2% and 82.8%, respectively, Sykora et al. [65] used a support vector machine classifier to categorize SIFT and SURF characteristics taken from 500 test photos. Preserving as much information about the original data as possible, PCA is a widely used technique for feature extraction and data dimensionality reduction that transforms high-dimensional data into low-dimensional data [54]. Frequently used for feature extraction and classification, LDA maximizes variability between categories and minimizes variability within them to transform high-dimensional data into low-dimensional data [66].

The low fraction of the gesture region in the image makes it difficult for global features to extract the relevant information from the hand region, which leads to poor image performance in the research of gesture image features. Local image features, on the other hand, are more numerous and stable, have lower intertexture correlation than global features, can partially prevent hand region occlusion, and are resilient to changes in illumination, rotation, and viewpoint. Thus, in order to attain higher identification rates, most researchers prefer to merge local features with global characteristics to enrich gesture feature information.

Accurate Gesture recognition is crucial for effective human-robot interaction (HRI). The ability of robots to precisely interpret and respond to human gestures can significantly enhance the usability and efficiency of collaborative tasks. In this context, various vision-based gesture recognition techniques have been developed, each offering different levels of accuracy which is shown in table 3. Understanding the comparative accuracy levels of these techniques is essential for designing robust and reliable HRI systems.

*Table 3. Comparison of Accuracy Levels of Different Vision-Based Gesture Recognition Techniques*

| Vision-Based Gesture Recognition Techniques | Accuracy 0Level | Reference |
|---|---|---|
| Convolutional Neural Networks (CNNs) | High | [67] |
| Hidden Markov Models (HMMs) | Moderate | [68] |
| Support Vector Machines (SVMs) | Moderate | [69] |
| Decision Trees | Moderate | [70] |
| Neural Networks | High | [71] |

## Machine Learning and Deep Learning Approaches:

Computers can generate models that adhere to the underlying principles of data in large datasets by utilizing machine learning. To control epidemic casualties, Tutsoy [72] created an enhanced prediction model under time-varying dynamics and a large number of uncertain elements. This multidimensional policy-making algorithm was based on artificial intelligence. Furthermore, a novel high-order, multidimensional, parametric, tightly linked suspicious-infected death model was proposed by Tutsoy [73]. To discover complicated associations and produce precise predictions, this model makes use of machine learning algorithms to learn from massive data sets, including epidemiological data, demographic data, and environmental factors. The challenge of gesture recognition can be solved by a variety of well-known machine learning classification techniques, including support vector machines, neural networks, conditional random fields, and k-nearest neighbor algorithms.

**Support Vector Clustering (SVC) algorithm:**

An algorithm for supervised machine learning is the Support Vector Clustering (SVC) algorithm. Kernel-based learning forms the basis of SVC. SVC employs both regression and classification methods to examine a pattern. Splitting a data set into two classes is the goal of using SVC in order to help the data identify a hyperplane [74]. This hyperplane serves as a binary classifier for us. In order to use these data points as support vectors, we must first search for the hyperplane's nearest data points. The creation of hyperplanes depends heavily on these support vectors [75]. The hyperplane's appearance and position are also influenced by support vectors.

**K-Nearest Neighbor (KNN) algorithm:**

The K-Nearest Neighbor (KNN) algorithm is a supervised machine-learning algorithm for gesture recognition. KNN is used for classification, where the categorization of a data point is based on the classification of its neighbor. In KNN, the closest neighbor is located using the Euclidean distance. The computation is done using multiple little distances, with the goal being to reach the smallest Euclidean distance [76]. Accuracy rises in tandem with the increase in k.

The Euclidian distance formula is typically applied. The KNN algorithm uses the threshold value, which is determined by averaging the k data points that are closest to it, to conduct classification. The performance is entirely dependent on the proximity to the closest neighbor, a similarity metric, and a threshold value. The number of neurons in the hidden layer is used to compute the hidden layer size and provide an accurate measurement. A solver is used to perform weight optimization.

**Logistic Regression algorithm:**

The goal of logistic regression is to illustrate the relationship between relevant variables and a discrete response variable by combining regression techniques. The way the response variable Y is continuous for straight regression is shown by the difference between standard straight regression and logistic regression. The reaction variable in logistic regression is discrete[77]. The selection of borders and suspicions demonstrates this divide. Because it uses a linear equation to predict values between 0 and 1, logistic regression is regarded as a predictive analytical algorithm. Its operation is based on probability. It converts the result into categorical numbers by using the activation function as sigmoid.

**Naïve Bayes algorithm:**

One common AI and information mining technique is classification. Different approaches may be chosen to carry out the classification job depending on the number of target categories that are used to categorize an informational index. Choice trees and backing vector machines are commonly used for paired classifications; however, both approaches rely on the requirement that there be no more than two goal classifications. Their inflexible constraint makes it challenging to accept them as fitting a broad variety of authentic classification operations, where the number of target classifications is typically greater than two. For broad classification assumptions, the Naive Bayes Classifier makes more sense. Naive Bayes classifiers have been used in a respectable number of real and useful applications, such as illness orders, client acknowledgment evaluations, and climate forecast services. Nonetheless, an even arrangement is preprocessed from the structure of an informational collection within the problem area [77]. The validity of fitting a fresh piece of data into each categorization can then be ascertained using this numerical classifier. As a result, the classification that best fits this piece of data can be determined by looking at the classification with the highest reliability.

**Stochastic Gradient Descent (SGD)**

Stochastic Gradient Descent (SGD) is a simple yet effective optimization method for identifying the benefits of capacity constraints that reduce work expenses. It is applied to discriminative learning of linear classifiers under arcs of misfortune, including logistic regression and SVM. Because the coefficients are changed for every preparation occasion rather than at the conclusion of instances, it has been applied to massive scope datasets with great success. The worst outcomes will come from SGDC if the hyperparameter selection is incorrect.

**Convolutional Neural Network (CNN):**

CNN is well-known for its recognition capabilities and produces better results than other approaches, mainly because it can extract the required element esteems from the information picture and learn to distinguish between distinct examples by using many examples during training [78]. In any case, the speed at which computing equipment is developing has historically restricted its growth. The CNN network has expanded rapidly in recent times because to the advancement of semiconductor fabrication, which has resulted in faster computation speeds for illustration preparation units and a bottleneck in equipment handling speed [75]. Using CNN requires the following steps: first, a picture must be entered (interpreted as an array of pixels); second, processing and filtering must be done; and finally, results must be retrieved following categorization. To be employed in a layered architecture with many convolutional layers involving kernels (or filters) and a pooling operation, each model must first be trained and then evaluated.

**Random Forest:**

A sequence of decision trees is used by a random forest [79] classifier to label the sample data. There are several decision trees available to assess the algorithm's performance. Furthermore, the technique used to create each of these trees was designed with prediction in mind. Because of their many benefits, random forest methods are chosen over other state-of-the-art techniques. These benefits include good performance in large datasets, lack of overfitting, variable usability as both numerical and categorical, ease of application in multi-class environments, and fewer parameter requirements.

**Analysis Using XGBoost**

An XGBoost model was trained by parameter adjustment [79, 80]. The two steps that were used are as follows:

1. To assess the model's overall performance, training was done against a baseline model.
2. By adjusting its parameters, a second model was trained, and the outcomes were compared to the baseline model.

It can be applied to regression in addition to classification. Because of features like parallel, distributed, and cache-aware computing, it is faster than other cutting-edge techniques. Scalability and optimization are a couple of the other attributes.

An overview of numerous noteworthy works in the field of hand gesture recognition is provided in table 4. This research investigates a range of methods and approaches, from unique algorithms like spatial fuzzy matching (SFM) and deep learning methodologies to real-time systems employing embedded convolutional neural networks (CNN). Each study's major conclusions and contributions emphasize developments in a variety of fields, including device control, healthcare, and human-computer interaction. These developments have also improved dependability, speed, accuracy, and usability. Researchers and practitioners can learn a great deal about the many tools and techniques used for precise and efficient hand gesture identification by looking through these research papers.

*Table 4. Summary of Studies on Hand Gesture Recognition*

| Reference | Authors | Approach | Key Findings |
|---|---|---|---|
| [78] | Tam et al. | Real-time hand gesture recognition system using embedded CNN for classification. | Improved reliability and execution, minimized reaction times, focus on ease of use. |
| [74] | MF Wahid et al. | Utilized EMG for hand gesture recognition without training ML algorithms. | Results compared with KNN, NB, DA, SVM, and RF. Valuable for human-PC interaction and controlling devices. |
| [81] | Li et al. | Introduced SFM algorithm for fused gesture dataset and achieved high accuracy in static and dynamic hand gestures. | SFM runs on simple machines, achieved 94-100% accuracy in static and ~90% in dynamic gestures. |
| [82] | Lee and Tanaka | Focused on finger identification and tracking using Kinect and Depth-Sense for natural user interface. | Provided natural communication and interface with finger and hand gesture recognition. |
| [83] | Nogales et al. | Proposed systematic literature review of hand gesture recognition based on infrared data and AI algorithms. | Identified gaps for future research, applicable in various fields like medicine, robotics, etc. |
| [84] | Allard et al. | Used deep learning for hand gesture recognition with datasets from Myo armband and Nina Pro database. | Deep learning applied to process multiple features from large datasets, improving gesture recognition. |

The accuracy of Naïve Bayes, K-Nearest Neighbor (KNN), random forest, XGBoost, logistic regression, support vector classifier (SVC), stochastic gradient descent classifier (SGDC), and convolution neural networks (CNN) is shown in Figure 1. A measure of a model's accuracy in labeling a sample as positive is called precision. The capacity to identify positive samples is measured by recall, and the F1-Score is utilized to strike a balance between recall and precision [85]. A hyperplane is located using SVM, and a classification report for SVC is produced.

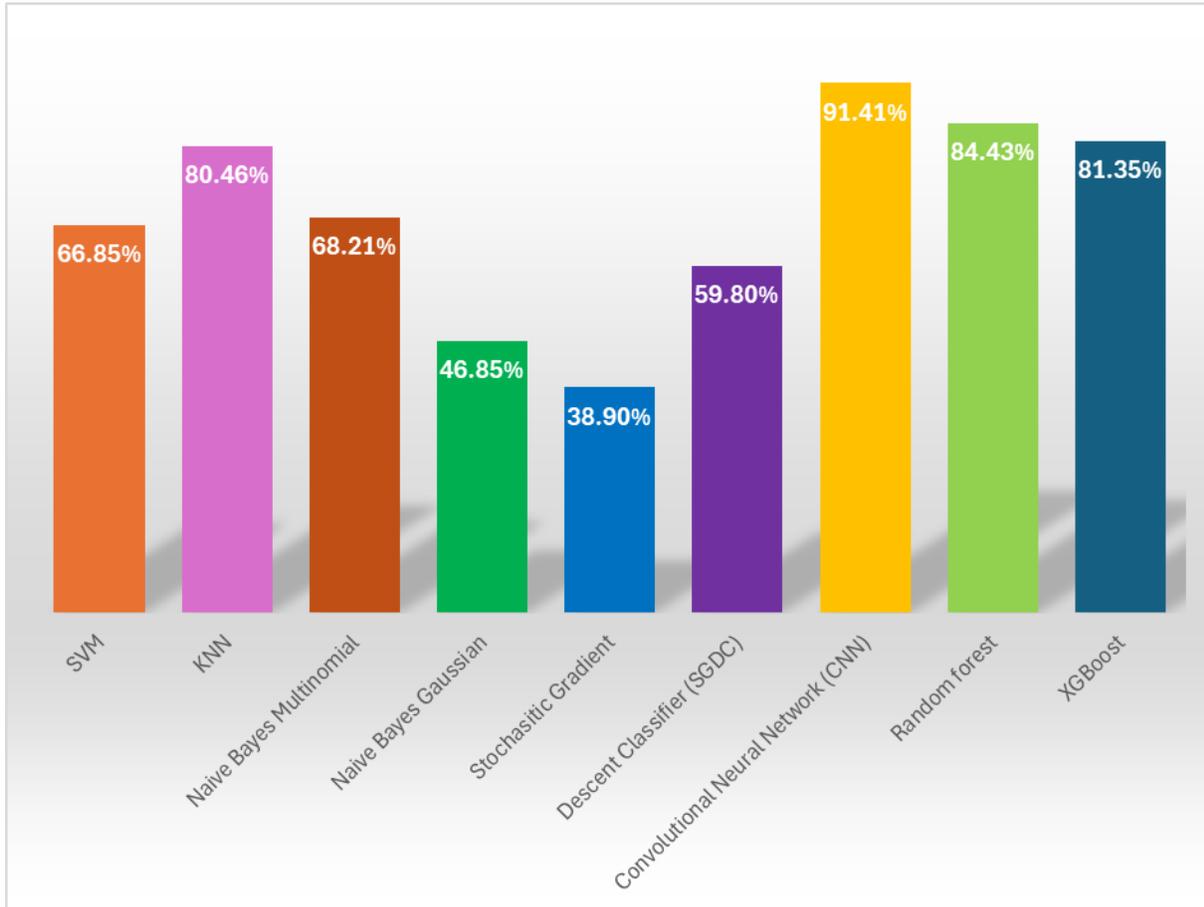

*Figure 1. Accuracy of selected Machine learning algorithm for gesture recognition.*

Applying SVC, KNN, Naïve Bayes Multinomial, Naïve Bayes Gaussian, Convolutional Neural Network (CNN), random forest, XGBoost, and Stochastic Gradient Descent Classifier (SGDC) to the sign MNIST dataset resulted in accuracies of 66.85, 80.46, 68.21, 46.85%, 38.9%, 59.80%, 91.41%, 84.43%, and 81.35%, respectively. Based on variables like accuracy, the classification report, and the confusion matrix, observations were made. Benefits of KNN [86] include robustness, simple debugging, and simplicity of execution. However, this method has certain drawbacks, including poor computational speed for large datasets and affectivity toward duplicity, as we discovered that the NB model had the least precision. To improve the findings, consider adjusting, smoothing, and preprocessing the dataset. Assumptions on the elements' freedom constitute a strong limitation, making the probability returns useless for ascertaining trustworthy outcomes. Now, CNN is a great way to get the finest classification results, and several modifications and enhancements are also being suggested in various studies.

## Applications of Gesture Recognition in Robotics:

Applications for gesture recognition are numerous and include virtual reality, safe driving, healthcare, and device control. This section primarily focuses on vision-based hand motions that are recorded by RGB-D and monocular cameras for use in human-robot interaction. The following is a list of the primary uses for gesture recognition technologies.

**Device control:** Gestures may be employed as well to operate intelligent robots. Millions of homes will gradually see home robots or smart home appliances as artificial intelligence advances and gesture control will become more comfortable for users than standard button or touchscreen input. uses is a startup that creates hardware and software to allow Smart TVs to identify gestures and finger movements. Gestoo's artificial intelligence platform allows touchless management of music and lighting systems through the use of gesture recognition technologies. One gesture can be used to activate numerous communication channels with Gestoo, which allows gestures to be designed and assigned from a smartphone or other device.

**Healthcare:** Individuals and equipment can create a lot of noise in chaotic emergency rooms and operation rooms. Voice commands are less successful in such a setting than hand gestures. The sharp distinctions between sterile and nonsterile environments rule out touchscreens as well. However, Microsoft has shown how gesture recognition technology can be used to view photos and information during surgery or other procedures. Using basic gestures instead of scrubbing, doctors may view MRI, CT, and other images with Gesture, a gesture control solution for medical devices. By minimizing the amount of time that nurses and doctors touch patients, this touch-free engagement lowers the possibility of cross-contamination.

**Awareness of sign language:** Sign language is the main communication tool for people with hearing impairments, however it can be challenging for non-trained individuals to comprehend sign language. Sign recognition technology for sign language cognition will significantly improve the communication skills of the deaf and other people.

**Virtual reality:** Gesture detection improves user immersion and experience by enabling more natural interactions and control over virtual reality situations. In addition to demonstrating improved gesture recognition software in 2016, Leap Motion enabled users to monitor motions in virtual reality while manipulating computers. Using a smartphone camera (on Android and iOS), Mano Motion's hand-tracking software can identify 3D movements and be used in AR and VR settings. Robotics, consumer electronics, gaming, and Internet of Things devices are some of the use cases for this technology.

## Challenges and Future Directions:

Deep learning is a benign gesture recognition accelerator, and as artificial intelligence advances, gesture recognition systems will get more precise and reliable. Future gesture recognition technologies will also be more varied and useful in a wider range of industries, including healthcare, education, entertainment, and so on, giving users greater convenience and innovation. With the progression of deep learning and artificial intelligence, the field of gesture recognition is poised to achieve greater levels of intelligence. Through training, models can become adept at identifying intricate movements with minimal user input, resulting in more sophisticated and intuitive gesture detection capabilities. In the future, advanced technology for recognizing gestures will have the ability to analyze and interpret vast quantities of movements in real time. This advancement will open up new possibilities for implementing gesture recognition in various fields such as virtual reality, gaming, medicine, and more. As the applications of gesture recognition continue to expand, the reliability of this technology becomes increasingly important. To ensure consistent and dependable performance across different environments, extensive testing and validation will be essential for future gesture recognition technologies. Advancements in computer vision and sensor technology will improve the accuracy of gesture detection. Higher-resolution cameras and more sensitive sensors, for instance, can record more minute hand movements, increasing the precision of gesture recognition. The gesture recognition technologies of the future will be more customized and able to adjust to the varied gesture habits and preferences of users. Users might be able to program individual gestures, for instance, to carry out a certain task or function.

## Conclusion:

This work investigates how to combine machine learning and deep learning algorithms with feature recognition approaches to transform collaboration and human-robot interaction (HRI) via effective gesture recognition. The innovations under examination improve reliability, safety, and smooth interaction between people and robots. The study also conducts a thorough analysis of numerous gesture recognition methodologies, covering applications, challenges, and potential prospects in the HRI industry. The study's overall findings demonstrate the potential for enhancing human-robot interaction (HRI) by combining sophisticated algorithms with feature recognition techniques. This will facilitate improved human-robot collaboration while tackling important issues and examining novel possibilities for research in this fascinating area of study.